\documentclass[runningheads]{llncs}
\usepackage{soul}
\usepackage{url}
\usepackage[hidelinks]{hyperref}
\usepackage[utf8]{inputenc}
\usepackage[small]{caption}
\usepackage{graphicx}
\usepackage{amsmath}
\usepackage{booktabs}
\usepackage{algorithm}
\usepackage{algorithmic}
\usepackage{multirow}
\usepackage{graphicx}
\usepackage{latexsym}
\usepackage{times}
\usepackage{xcolor}
\usepackage{colortbl}
\usepackage{graphicx}
\usepackage{enumitem}
\usepackage{algorithm}
\usepackage{algorithmic}
\usepackage{makecell}

\usepackage{amsthm}

\usepackage[capitalize]{cleveref}
\crefname{section}{Sec.}{Secs.}
\Crefname{section}{Section}{Sections}
\Crefname{table}{Table}{Tables}
\crefname{table}{Tab.}{Tabs.}

\def\model{MH-pFLGB}
\def\modelf{Model Heterogeneous personalized Federated Learning via Global Bypass}

\begin{document}
\title{MH-pFLGB: Model Heterogeneous personalized Federated Learning via Global Bypass for Medical Image Analysis}
%
%

\author{Luyuan Xie\inst{1,2}\thanks{Equal Contribution.}\and Manqing Lin\inst{1\star} \and ChenMing Xu\inst{1,2}\and Tianyu Luan \inst{3} \and Zhipeng Zeng\inst{1,2} \and Wenjun Qian\inst{1,2}\and Cong Li\inst{1,2}\thanks{Corresponding author: Cong Li, li.cong@pku.edu.cn; Zhonghai Wu, wuzh@pku.edu.cn}\and Yuejian Fang\inst{1,2}\and Qingni Shen\inst{1,2}\and Zhonghai Wu\inst{1,2\star\star}}

\institute{$^{1}$School of Software and Microelectronics, Peking University, Beijing, China \\$^{2}$National Engineering Research Center for Software Engineering, Peking University, Beijing 100871, China \\$^{3}$State University of New York at Buffalo }

%
%
\maketitle              
\begin{abstract}
In the evolving application of medical artificial intelligence, federated learning is notable for its ability to protect training data privacy. 
Federated learning facilitates collaborative model development without the need to share local data from healthcare institutions. Yet, the statistical and system heterogeneity among these institutions poses substantial challenges, which affects the effectiveness of federated learning and hampers the exchange of information between clients.
To address these issues, we introduce a novel approach, \model{}, which employs a global bypass strategy to mitigate the reliance on public datasets and navigate the complexities of non-IID data distributions. Our method enhances traditional federated learning by integrating a global bypass model, which would share the information among the clients, but also serves as part of the network to enhance the performance on each client. Additionally, \model{} provides a feature fusion module to better combine the local and global features. We validate \model{}'s effectiveness and adaptability through extensive testing on different medical tasks, demonstrating superior performance compared to existing state-of-the-art methods.

\end{abstract}
\section{Introduction}
In the field of medical images, federated learning \cite{fedavg} has emerged as a key technique for its ability to protect the privacy of training datasets. This approach allows for the collaborative development of a unified global model, eliminating the need to directly share local data from individual healthcare facilities. However, the application of federated learning in healthcare faces challenges such as statistical heterogeneity \cite{knnper}, due to the diverse and non-uniformly distributed (non-IID) data across different institutions, and system heterogeneity \cite{knnper}, due to the unique architecture of local models by each institution. These challenges compromise the efficiency of federated learning and hinder the seamless exchange of information between client models. Addressing the issues of statistical and system heterogeneity presents a critical and impactful challenge in the application of federated learning within healthcare facilities.

Previous works only focused on statistical heterogeneity and proposed personalized federated learning methods \cite{fedsm2022,clustered1,apfl,lcfed,li2021ditto,fedrep,lg-fedavg}. Compared to traditional single model settings \cite{fedavg,karimireddy2020scaffold,fedprox}, personalized federated learning allows each client to learn their own model, effectively alleviating the problem of statistic heterogeneity. However, these methods still require models with the same structure for each client. Recent works including FedMD~\cite{DBLP:journals/corr/abs-1910-03581}, FedDF~\cite{NEURIPS2020_18df51b9}, DS-pFL \cite{9392310} and KT-pFL~\cite{NEURIPS2021_5383c731} tackle statistic and system heterogeneity in federated learning by sharing soft predictions among clients. 
These approaches have advanced the field by addressing heterogeneity issue, but depend heavily on public datasets for generating these soft predictions. However, collecting the medical dataset for public usage would involve a certain level of privacy requirements and complex censoring processes. Besides, the extensive size of public datasets would largely increase the computational cost, thus limiting the application of these techniques. All of these problems would significantly raise the cost of deploying those methods.


To eliminate the reliance on public datasets, we propose a global bypass strategy to address the challenges of heterogeneous models under the distribution of non-IID data. Unlike traditional approaches that rely on soft prediction generated from public datasets, our method adds a global bypass model to the local clients to share the information among the clients and help the local clients. In each client, the global bypass would not only learn the information from local data, but also help the previous local network to make its prediction. In the server, we aggregate the global bypass to share the information among each client. Additionally, we design the global bypass to be small so the computational cost is less than what would be required for local training on a public dataset.


Specifically, we propose framework \modelf{} (\model{}). Our global bypass consists of a body and a head module. The body is a light-weighted encoder for feature extraction and the head is a small module designed to fit the outputs of different tasks. To better fuse the information from the local model and global bypass model, we designed a fusion module named features weighted fusion to fusion the features from the body of the local and global model. The fusion is based on allowing models to learn how to better select weights for global and local features. This design would better utilize global knowledge and integrate it with local features, so that it can improve the performance of local models for each client.

Our contributions are summarized as follows:
\begin{itemize}[noitemsep,topsep=0pt]
\item We introduce a novel personalized federated learning approach for dealing with heterogeneous models named \model{}. This approach leverages a global bypass mechanism that obviates the need for public medical datasets, thereby reducing the additional burdens associated with local training.
\item We design a global bypass model to transfer information among different clients and enhance the result of each local client. Additionally, we integrate a feature fusion module to more effectively combine features from the local model and the global bypass.
\item We demonstrate the efficacy and versatility of our \model{} through rigorous testing on a variety of medical tasks, such as image classification and image segmentation. Our method surpasses current state-of-the-art results in all these areas, underscoring its potential and adaptability across a broad range of medical applications.

\end{itemize}

\begin{figure}[t]
\includegraphics[width=\textwidth]{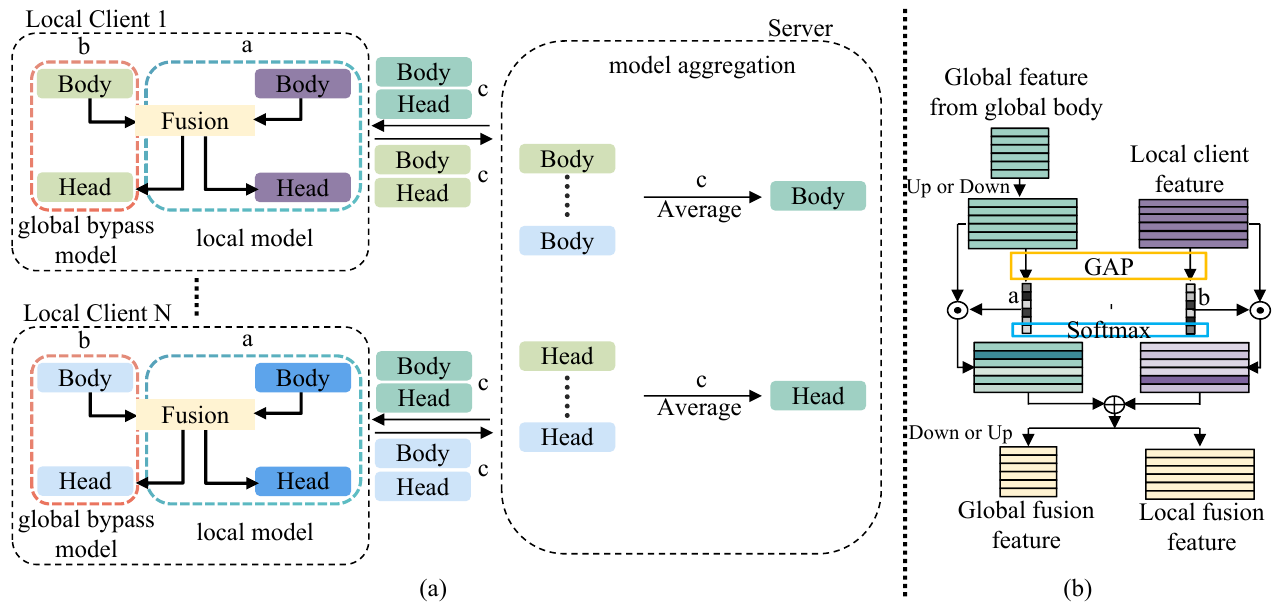}
\caption{(a) Overview of our proposed \model{} framework. Each training process consists of 3 steps. From a to c: \textbf{a.} Local model training. \textbf{b.} Global bypass model training. \textbf{c.} Upload, aggregation, and download. (b) Features weighted fusion. More details can be found in  Section 2.1 and Section 2.2.} \label{fig:pipeline}
\end{figure}
\section{Methods}

\subsection{Pipeline}

The pipeline of \model{} is shown in \cref{fig:pipeline} (a). 
Each local client consists of an architecture heterogeneous local model and a global bypass model that shares the same network architecture among other clients. The local body extracts personalized features from local clients, while the global body shares learned parameters among clients.
Both local and global models are divided into a body model to extract features, and a head model to generate the network output using the features. 
Our training process consists of 3 steps: \textbf{a.} Local model training, \textbf{b.}  Global bypass model training, and \textbf{c.} Global aggregation. We will explain the details of those steps in the following.

\textbf{Local model training.} 
In the local model training stage, 
the local model learns from both the local dataset and the global insights provided by the bypass model.
At this stage, we freeze the global bypass model and only train the local model. For client $i$, its local model training loss function $\mathcal{L}_{\textit{loc},i}$ is:
\begin{equation} 
{\mathcal{L}_{\textit{loc},i}} = {\lambda^l_{\textit{loc}}}{\mathcal{L}^l_{\textit{loc}}(\hat{y}^l_{i},y_{i})}  + {\lambda^g_{\textit{loc}}}{\mathcal{L}^g_{\textit{loc}}(\hat{y}^g_{i},y_{i})},
\label{1}
\end{equation}
\noindent
where 
$\hat{y}^l_{i}$ and $\hat{y}^g_{i}$ are the predictions from the local head and global head. 
$\mathcal{L}^l_{\textit{loc}}$ and $\mathcal{L}^g_{\textit{loc}}$  represent the loss functions for the local and global model output, respectively. $\lambda^l_{\textit{loc}}$ and $\lambda^g_{\textit{loc}}$ are their corresponding weights. $y_{i}$ is the label of input data $x_{i}$. Note that, even though the global bypass is fixed, we still calculate the loss function on its output to maintain generalizability when training the local model.

\textbf{Global bypass model training.} 
During the global model training phase, we freeze the local model and fine-tune the global bypass model. This enables the body of the global model to learn the information 
from each client.
The loss function $\mathcal{L}_{\textit{glob},i}$ is represented as:
\begin{equation} 
{\mathcal{L}_{\textit{glob},i}} = {\lambda^g_{\textit{glob}}}{\mathcal{L}^g_{\textit{glob}}(\hat{y}^g_{i},y_{i})}  + {\lambda^{l}_{\textit{glob}}}{\mathcal{L}^{l}_{\textit{glob}}(\hat{y}^l_{i},y_{i})}.
\end{equation} 
\noindent
$\mathcal{L}^g_{\textit{glob}}$ and $\mathcal{L}^{l}_{\textit{glob}}$ represent the loss function for training the global and local model at the global training stage, respectively. $\lambda^g_{\textit{glob}}$ and $\lambda^{l}_{\textit{glob}}$ are their corresponding weights. Other variables are defined the same as in eq.(1). Similar to the local training stage, local loss function $\mathcal{L}^{l}_{\textit{glob}}$ is designed to avoid client drift.

\textbf{Global Aggregation.}  
As the global model is uploaded to the server, the global aggregation process aggregates the model parameters, with distinct processes for both the body and head of the global model. This aggregation employs weight averaging, as outlined in~\cite{fedavg}. Finally, the aggregated model is downloaded and distributed for the next round of training. 

At the inference stage, we fuse global and local features, and the fused features output prediction results through the local head. The global head only participates in the training stage and does not participate in the inference stage. This is similar to adding a regularization term during local training, effectively preventing overfitting of the local model.

\subsection{Features Weighted Fusion}

In order to better fuse global and local features, we propose a new feature method named Feature Weighted Fusion. 
As shown in \cref{fig:pipeline}(b), the feature from global body $x_g$ ensures that the dimension is the same as local client feature $x_l$ through upsampling or downsampling as 
\begin{equation} 
\begin{array}{c}
     \hat{x}_g = f_{\textit{up}}(x_{g}) \ or \ f_{\textit{down}}(x_g),
\end{array}
\end{equation} 
where $f_{\textit{up}}(\cdot)$ and $f_{\textit{down}}(\cdot)$ respectively represent upsampling or downsampling operations. Specifically, in the classification task, we use $1 \times 1$ convolution for both upsampling and downsampling. In the segmentation task, we adopt deconvolution for upsampling and convolution for downsampling, along with a global average pooling operation on the results. Having $\hat{x}_g$, a Softmax operator is applied on $\hat{x}_g$ and $x_l$ ’s channel-wise digits:
\begin{equation} 
\begin{array}{c}
     a_i = \frac{\exp(\hat{x}_{g,i})}{\exp(\hat{x}_{g,i}) + \exp(x_{l,i})}; \ b_i = \frac{\exp(x_{l,i})}{\exp(\hat{x}_{g,i}) + \exp(x_{l,i})},\ 0 < i \leq C,
\end{array}
\end{equation} 
\noindent
where $C$ is the the dimension of $\hat{x}_g$ and $x_l$. $a$ and $b$ are the calculated weights. The local fusion feature $x_{\textit{lf}}$ is obtained by multiplying and adding the corresponding weights and features. 
\begin{equation} 
\begin{array}{c}
     x_{\textit{lf},i} = a_i \hat{x}_{g,i} + b_i x_{l,i}, \ 0 < i \leq C,
\end{array}
\end{equation} 
where $x_{\textit{lf},i}$ is the $i$-th element of $x_{\textit{lf}}$. $x_{\textit{lf}}$ obtains the global fusion feature $x_{\textit{gf}}$ for global head through downsampling or upsampling as:
\begin{equation} 
\begin{array}{c}
     x_{\textit{gf}} = f_{\textit{down}}(x_{\textit{lf}}) \ or \ f_{\textit{up}}(x_{\textit{lf}}).
\end{array}
\end{equation} 


\begin{table*}[t]
  \centering
  \caption{The results of classification task in different resolutions. The x2↓, x4↓, and x8↓ are downsampling half, quarter, and eighth of high-resolution images. We evaluate ACC and MF1 results on BreaKHis dataset. The larger the better. \textbf{Bold} number means the best. 
  Only local training, FedMD, FedDF, pFedDF, DS-PFL, KT-PFL, and \model{} use heterogeneous models in each client. The four client models are set to ResNet$\lbrace 17,11,8,5 \rbrace$, respectively. Other methods use the unified model settings (ResNet17). \model{} achieves the best performance.}
  \resizebox{0.90\linewidth}{!}{
    \begin{tabular}{c|cc|cc|cc|cc|cc}
    \hline
    \multicolumn{1}{c|}{\multirow{2}[1]{*}{Method}} & \multicolumn{2}{c|}{high-resolution} & \multicolumn{2}{c|}{x2↓} & \multicolumn{2}{c|}{x4↓} & \multicolumn{2}{c|}{x8↓} & \multicolumn{2}{c}{Average} \\
     \cline{2-11} 
    \multicolumn{1}{c|}{} & ACC$\uparrow$   & MF1$\uparrow$   & ACC$\uparrow$   & MF1$\uparrow$   & ACC$\uparrow$  & MF1$\uparrow$   & ACC$\uparrow$   & MF1$\uparrow$   & ACC$\uparrow$   & MF1$\uparrow$ \\
    \hline
     Only Local Training & 0.7891  & 0.7319  & 0.8027  & 0.7461  & 0.7538  & 0.6852  & 0.6956  & 0.5867  & 0.7603  & 0.6875  \\
   \hline
    FedAvg \cite{fedavg} & 0.7406  & 0.6425  & 0.7908  & 0.7405  & 0.6892  & 0.6031  & 0.5774  & 0.4681  & 0.6995  & 0.6136  \\
    FedAvg+FT & 0.7749 & 0.7218 & 0.8124 & 0.7511 & 0.7327 & 0.6628 & 0.6234 & 0.5073 & 0.7359  & 0.6608 \\
    SCAFFOLD \cite{karimireddy2020scaffold}&0.7442  & 0.6512  & 0.8097  & 0.7533  & 0.6725  & 0.5963  & 0.5866  & 0.4732  & 0.7033  & 0.6185  \\
    SCAFFOLD+FT & 0.7761 & 0.7229 & 0.8237 & 0.7709 & 0.7523 & 0.6872 & 0.6142 & 0.5005 & 0.7416  & 0.6704 \\
    FedProx \cite{fedprox}& 0.7354  & 0.6386  & 0.7873  & 0.7421  & 0.6944  & 0.6107  & 0.5821  & 0.4687  & 0.6998  & 0.6150 \\
    FedProx+FT &0.7827 & 0.732 & 0.8055 & 0.7549 & 0.7548 & 0.6811 & 0.6071 & 0.4829 & 0.7375  & 0.6627 \\
    \hline
     Ditto \cite{li2021ditto}& 0.7304  & 0.6221  & 0.7661  & 0.6482  & 0.6065  & 0.5022  & 0.5931  & 0.4741  & 0.6740  & 0.5617  \\
     APFL \cite{apfl}& 0.7444  & 0.6568  & 0.7992  & 0.7355  & 0.6227  & 0.5229  & 0.6133  & 0.4986  & 0.6949  & 0.6035  \\
     FedRep \cite{fedrep}& 0.7991  & 0.7618  & 0.8229  & 0.7697  & 0.7762  & 0.7182  & 0.6328  & 0.5091  & 0.7578  & 0.6897 \\
     LG-FedAvg \cite{lg-fedavg}& 0.7972  & 0.7523  & 0.5655  & 0.4397  & 0.6131  & 0.5080  & 0.6080  & 0.4902  & 0.6460  & 0.5476  \\
    \hline
     FedMD \cite{DBLP:journals/corr/abs-1910-03581}& 0.7599  & 0.7083  & 0.8321  & 0.7829  & 0.7721  & 0.6893  & 0.6495  & 0.5439  & 0.7534  & 0.6811  \\
     FedDF \cite{NEURIPS2020_18df51b9}& 0.7661  & 0.7253  & 0.8132  & 0.7629  & 0.7826  & 0.7342  & 0.6627  & 0.5627  & 0.7562  & 0.6963    \\
     pFedDF \cite{NEURIPS2020_18df51b9}& 0.8233  & 0.7941  & 0.8369  & 0.7965  & 0.8121  & 0.7534  & 0.6843  & 0.6022  & 0.7892  & 0.7366  \\
     DS-pFL \cite{9392310}& 0.7842  & 0.7609  & 0.8334  & 0.7967  & 0.7782  & 0.7258  & 0.6327  & 0.5229  & 0.7571  & 0.7016 \\
      KT-pFL \cite{NEURIPS2021_5383c731}& 0.8424  & 0.8133  & 0.8441  & 0.8011  & 0.7801  & 0.7325  & 0.7032  & 0.6219  & 0.7925  & 0.7422  \\
    \hline
      \model{} (Ours)  & \textbf{0.8952}  & \textbf{0.8697}  & \textbf{0.8963}  & \textbf{0.8745}  & \textbf{0.8681}  & \textbf{0.8333}  & \textbf{0.7793}  & \textbf{0.7214}  & \textbf{0.8597}  & \textbf{0.8247}  \\
    \hline
    \end{tabular}%
  \label{tab:rc}%
   }
\end{table*}%

\section{Experiments Setup}

\textbf{Task and Dataset.} We verify the effectiveness of \model{} on 3 non-IID tasks. For \textbf{medical image classification (different resolution)} task, our experiments are conducted on the \textit{Breast Cancer Histopathological Image Database (BreaKHis)}~\cite{7312934}. We perform x2↓, x4↓, and x8↓ downsampling on the high-resolution images \cite{xie2023shisrcnet}. Each resolution of medical images is treated as a client. In this task, we employed ResNet $\{17,11,8,5\}$ as the local model of each client, respectively. For \textbf{medical image classification (different label distributions)} task, we employ 2 datasets, including a breast cancer classification dataset \textit{BreaKHis} (color images) and an Optical Coherence Tomography (OCT) disease classification dataset \textit{OCT2017} (grayscale images) \cite{kermany2018identifying}. We design eight clients, each corresponding to a distinct heterogeneous model. 
These models include ResNet \cite{he2015deep}, ShuffleNetV2 \cite{ma2018shufflenet}, ResNeXt \cite{xie2017aggregated}, SqueezeNet \cite{iandola2016squeezenet}, SENet \cite{hu2018squeeze}, MobileNetV2 \cite{sandler2018mobilenetv2}, DenseNet \cite{huang2017densely}, and VGG \cite{simonyan2014very}. For \textbf{medical image segmentation} task, we use four datasets for polyp segmentation. They are ColonDB \cite{Tajbakhsh2015}, ETIS \cite{Zhou2014}, ClinicDB \cite{Bernal2015} and Kvasir-SEG \cite{Jha2020}. Each center's dataset treated as a separate client. Each client utilized a specific model, including Unet++ \cite{zhou2019unet++}, FCN \cite{long2015fully}, Unet \cite{ronneberger2015u}, and Res-Unet \cite{diakogiannis2020resunet}.

\begin{table*}[ht]
  \centering
  \caption{The results of Image Classification Task with Different Label Distributions. This task includes breast cancer classification and OCT disease classification. We evaluate ACC and MF1 result in this task. The larger the better. \textbf{Bold} number means the best. MH-pFLID has the best performance.}
  \resizebox{1\linewidth}{!}{
    \begin{tabular}{c|cc|cc|cc|cc|cc|cc|cc|cc|cc}
    \hline
    \multicolumn{19}{c}{Breast cancer classification} \\
    \hline
    \multirow{2}[1]{*}{Method} & \multicolumn{2}{c}{ResNet} & \multicolumn{2}{c}{shufflenetv2} & \multicolumn{2}{c}{ResNeXt} & \multicolumn{2}{c}{squeezeNet} & \multicolumn{2}{c}{SENet} & \multicolumn{2}{c}{MobileNet} & \multicolumn{2}{c}{DenseNet} & \multicolumn{2}{c}{VGG} & \multicolumn{2}{c}{Average} \\
\cline{2-19}          & ACC$\uparrow$   & MF1$\uparrow$   & ACC$\uparrow$   & MF1$\uparrow$   & ACC$\uparrow$   & MF1 $\uparrow$  & ACC$\uparrow$   & MF1$\uparrow$   & ACC$\uparrow$   & MF1$\uparrow$   & ACC$\uparrow$   & MF1$\uparrow$   & ACC$\uparrow$   & MF1$\uparrow$   & ACC$\uparrow$   & MF1$\uparrow$   & ACC$\uparrow$   & MF1$\uparrow$ \\
    \hline
    Only Local Training & 0.59  & 0.455 & 0.845 & 0.8412 & 0.665 & 0.5519 & 0.84  & 0.7919 & 0.875 & 0.849 & 0.755 & 0.5752 & 0.855 & 0.6884 & 0.875 & 0.8515 & 0.7875  & 0.7005  \\
    \hline
    FedMD \cite{DBLP:journals/corr/abs-1910-03581}& 0.692 & 0.5721 & 0.823 & 0.8027 & 0.704 & 0.6087 & 0.875 & 0.8544 & 0.907 & 0.8745 & 0.762 & 0.6627 & 0.835 & 0.6493 & 0.842 & 0.8001 & 0.8050  & 0.7281  \\
    FedDF \cite{NEURIPS2020_18df51b9}& 0.721 & 0.5949 & 0.817 & 0.8094 & 0.723 & 0.6221 & 0.893 & 0.8735 & 0.935 & 0.9021 & 0.757 & 0.6609 & 0.847 & 0.6819 & 0.833 & 0.7826 & 0.8158  & 0.7409  \\
    pFedDF \cite{NEURIPS2020_18df51b9}& 0.755 & 0.6536 & 0.853 & 0.8256 & 0.741 & 0.6237 & 0.894 & 0.8742 & 0.935 & 0.9021 & 0.796 & 0.7219 & 0.879 & 0.7095 & 0.874 & 0.8521 & 0.8409  & 0.7703  \\
    DS-pFL \cite{9392310}& 0.715 & 0.6099 & 0.792 & 0.7734 & 0.765 & 0.6547 & 0.899 & 0.8792 & 0.935 & 0.9021 & 0.794 & 0.7331 & 0.853 & 0.6691 & 0.851 & 0.8266 & 0.8255  & 0.7560  \\
    KT-pFL \cite{NEURIPS2021_5383c731}& 0.765 & 0.6733 & 0.87  & 0.8331 & 0.755 & 0.6432 & 0.885 & 0.8621 & 0.935 & 0.9021 & 0.78  & 0.6931 & 0.865 & 0.6819 & 0.905 & 0.9023 & 0.8450  & 0.7739  \\
    \hline
    \model{} (Ours)  & \textbf{0.830}  & \textbf{0.7076} & \textbf{0.945}  & \textbf{0.9399} & \textbf{0.820}  & \textbf{0.7796} & \textbf{0.975} & \textbf{0.9608} & \textbf{0.962}  & \textbf{0.9452} & \textbf{0.821} & \textbf{0.7080} & \textbf{0.893} & \textbf{0.7015} & \textbf{0.995}  & \textbf{0.9928} & \textbf{0.9038}  & \textbf{0.8419}  \\
    \hline
    \multicolumn{19}{c}{OCT disease classification } \\
    \hline
    \multirow{2}[1]{*}{Method} & \multicolumn{2}{c}{ResNet} & \multicolumn{2}{c}{shufflenetv2} & \multicolumn{2}{c}{ResNeXt} & \multicolumn{2}{c}{squeezeNet} & \multicolumn{2}{c}{SENet} & \multicolumn{2}{c}{MobileNet} & \multicolumn{2}{c}{DenseNet} & \multicolumn{2}{c}{VGG} & \multicolumn{2}{c}{Average} \\
\cline{2-19}          & ACC$\uparrow$   & MF1$\uparrow$   & ACC$\uparrow$   & MF1$\uparrow$   & ACC$\uparrow$   & MF1$\uparrow$   & ACC$\uparrow$   & MF1$\uparrow$   & ACC $\uparrow$  & MF1$\uparrow$   & ACC$\uparrow$   & MF1$\uparrow$   & ACC$\uparrow$   & MF1$\uparrow$   & ACC$\uparrow$   & MF1$\uparrow$   & ACC$\uparrow$   & MF1$\uparrow$ \\
    \hline
    Only Local Training & 0.9162 & 0.9099 & 0.8922 & 0.8918 & 0.8694 & 0.8253 & 0.8472 & 0.8361 & 0.9388 & 0.9311 & 0.914 & 0.7236 & 0.9054 & 0.9   & 0.9262 & 0.9077 & 0.9012  & 0.8657  \\
    \hline
    FedMD \cite{DBLP:journals/corr/abs-1910-03581}& 0.8828 & 0.8349 & 0.8856 & 0.8531 & 0.8246 & 0.7822 & 0.8254 & 0.8021 & 0.8552 & 0.8321 & 0.9254 & 0.7542 & 0.9254 & 0.9119 & 0.9552 & 0.9293 & 0.8850  & 0.8375  \\
    FedDF \cite{NEURIPS2020_18df51b9}& 0.854 & 0.8229 & 0.913 & 0.8936 & 0.865 & 0.8241 & 0.8054 & 0.7749 & 0.8926 & 0.8733 & 0.9178 & 0.7361 & 0.8958 & 0.8831 & 0.963 & 0.9308 & 0.8883  & 0.8424  \\
    pFedDF \cite{NEURIPS2020_18df51b9}& 0.9364 & 0.9152 & 0.92  & 0.913 & 0.881 & 0.8327 & 0.863 & 0.8239 & 0.941 & 0.8952 & 0.931 & 0.7249 & 0.897 & 0.8829 & 0.961 & 0.9234 & 0.9163  & 0.8639  \\
    DS-pFL \cite{9392310}& 0.8432 & 0.8079 & 0.864 & 0.8604 & 0.874 & 0.8356 & 0.835 & 0.7449 & 0.8874 & 0.8821 & 0.8998 & 0.7532 & 0.8592 & 0.8264 & 0.8814 & 0.8731 & 0.8680  & 0.8230  \\
    KT-pFL \cite{NEURIPS2021_5383c731}& 0.9532 & 0.9392 & 0.965 & 0.963 & 0.8594 & 0.8466 & 0.9136 & 0.9067 & 0.955 & 0.943 & 0.9622 & 0.8099 & 0.9038 & 0.8794 & 0.927 & 0.9022 & 0.9299  & 0.8988  \\
    \hline
    \model{} (Ours)  & \textbf{0.9634} & \textbf{0.9507} & \textbf{0.9950} & \textbf{0.9864} & \textbf{0.8824} & \textbf{0.8539} & \textbf{0.9672} & \textbf{0.9573} & \textbf{0.9714} & \textbf{0.9639} & \textbf{0.9674} & \textbf{0.8055} & \textbf{0.9088} & \textbf{0.9020} & \textbf{0.9496} & \textbf{0.9288} & \textbf{0.9506}  & \textbf{0.9186}  \\
    \hline
    \end{tabular}%
  \label{table_lc}%
  }
\end{table*}%

\begin{table}[t]
  \centering

\end{table}%

\textbf{Implementation Details.} \model{} adopts learning rates of $1\times10^{-4}$ and $1\times10^{-5}$ for the local model training and global model training stage, respectively. The batch size is set to 8. In experiments, all frameworks have a communication round of 100. Local training epochs are 5 (4 epochs in the first stage and 1 round in the second stage for \model{}). For classification, all of the loss functions are cross-entropy loss, and all of the loss functions are Dice loss for segmentation tasks. $\lambda^l_{\textit{local}}$ and $\lambda^g_{\textit{glob}}$ are set to 0.9. $\lambda^g_{\textit{loc}}$ and $\lambda^{l}_{\textit{glob}}$ are 0.1. The performance evaluation of the classification task is based on two metrics, accuracy (\textit{ACC}) and macro-averaged F1-score (\textit{MF1}), providing a comprehensive assessment of the model's robustness \cite{xie2024trls}. Additionally, \textit{Dice} is used to evaluate the segmentation task performance across frameworks \cite{fedsm2022}. Moreover, we implement \model{} using PyTorch 1.10 \cite{paszke2019pytorch} and train it on an NVIDIA GeForce RTX 3090 Ti GPU. We have included more baseline, datasets, training settings, and model structure details in the supplementary materials.

\begin{figure}[t]
\includegraphics[width=\textwidth]{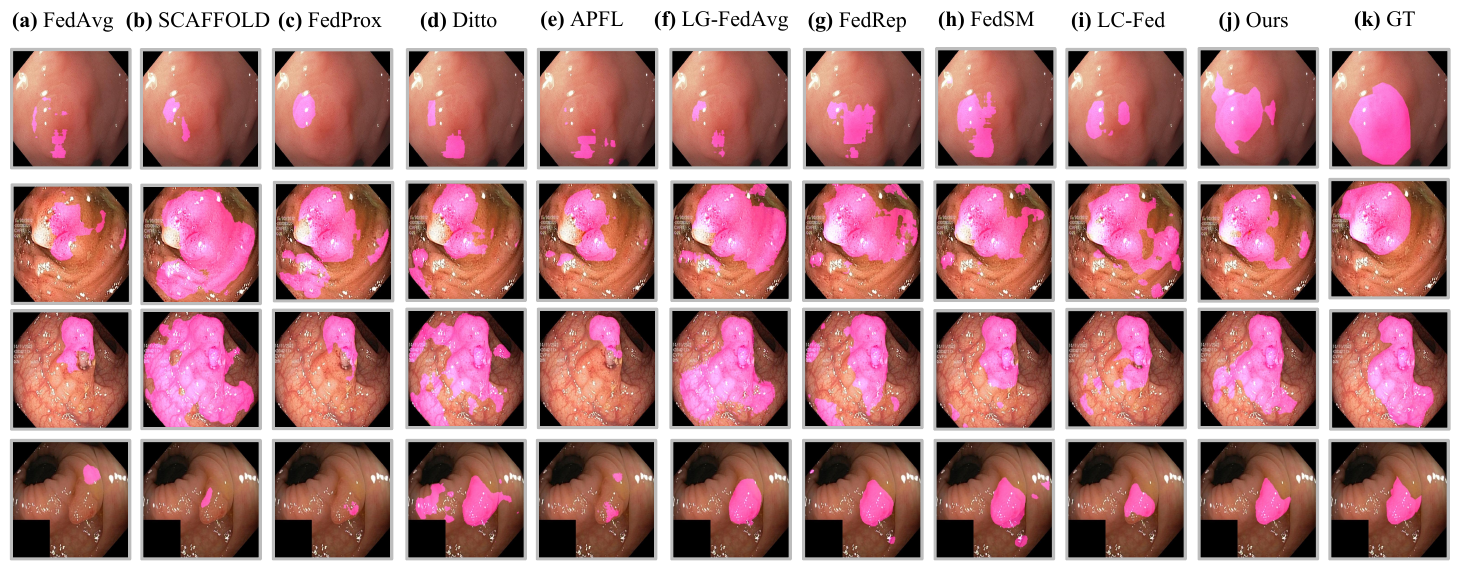}
\caption{Visualized comparison of Federated Learning in medical image segmentation. We randomly select four samples from different clients to form the visualization. (a-j)  Segmentation results by a model trained with FedAVG, SCAFFOLD, FedProx, Ditto, APFL, LG-FedAvg, FedRep, FedSM, LC-Fed, and our method \model{}; (k) Ground truths (denoted as ‘GT’).} 
\label{fig2}
\end{figure}

\section{Results and Discussion}
\subsection{Medical Image Classification (Different Resolutions)} 
In this task, we employ the ResNet family model to train breast cancer medical images under different resolutions. The higher the image resolution of this client, the deeper and more complex the model we adopt. In Table 1, compared to other federated learning frameworks, \model{} achieves the best performance. This indicates that \model{}, based on the global bypass model paradigm, effectively enables local heterogeneous models within the same family to fuse global knowledge, thereby enhancing the performance of local models. Furthermore, \model{} demonstrates a more significant advantage in terms of the MF1 metric, highlighting its ability to improve the robustness of local heterogeneous models.

\subsection{Medical Image Classification (Different Label Distributions)} 
In Table 2, the experimental results for the medical image classification task with different label distributions, where each client uses heterogeneous models, show that \model{} achieves the optimal results. This demonstrates that, compared to heterogeneous federated learning methods based on soft predictions, the global bypass model approach of \model{} has advantages. It can more effectively utilize knowledge from other clients to guide local client learning. Compared to only local training, \model{} enhances the local performance of each heterogeneous model. This indicates that our proposed feature weighted fusion method fuses global and local features well, thereby improving the performance of local models.

\subsection{Medical Image Segmentation}
 We validate the effectiveness of \model{} in medical image segmentation tasks. Table 3 presents the results of previous federated learning frameworks in the segmentation task, demonstrating that \model{} achieves the best outcomes. This indicates that our framework can effectively fuse global features and local heterogeneous model features from various clients, thus performing well in various downstream tasks. Meanwhile, the visualization results in Fig.~\ref{fig2} show that the segmentation results of \model{} are closer to ground truth.

\begin{table}[t]
\begin{minipage}[h]{0.45\textwidth}
\centering
\captionof{table}{ 
For the medical image segmentation task, we evaluate the \textit{Dice} result on Polyp dataset. The larger the better. Row ``Only Local Training" and ``\model{}" use heterogeneous models in each client. The four client models are set to Unet++, FCN, Unet, and ResUnet, respectively. For other methods, their clients use the unified model settings (Unet). \model{} achieves the best segmentation results.}
\resizebox{1\linewidth}{!}{
\begin{tabular}{c|c|c|c|c|c}
\hline
Methods & Client1 & Client2 & Client3 & Client4 & Average \\
\hline
{FedAvg} \cite{fedavg}& 0.5249  & 0.4205  & 0.5676  & 0.5500  & 0.5158  \\
FedAvg+FT & 0.6047  & 0.4762  & 0.7513  & 0.6681  & 0.6251  \\
 SCAFFOLD \cite{karimireddy2020scaffold}& 0.5244  & 0.3591  & 0.5935  & 0.5713  & 0.5121  \\
SCAFFOLD+FT & 0.5937  & 0.4312  & 0.8231  & 0.7208  & 0.6422  \\
 FedProx \cite{fedprox}& 0.5529  & 0.4674  & 0.5403  & 0.6301  & 0.5477  \\
 FedProx+FT & 0.7441  & 0.5701  & 0.7438  & 0.6402  & 0.6746  \\
\hline
Ditto \cite{li2021ditto}& 0.5720  & 0.4644  & 0.6648  & 0.6416  & 0.5857  \\
 APFL  \cite{apfl}& 0.6120  & 0.5095  & 0.6333  & 0.5892  & 0.5860  \\
LG-FedAvg \cite{lg-fedavg}& 0.6053  & 0.5062  & 0.7371  & 0.5596  & 0.6021  \\
FedRep \cite{fedrep}& 0.5809  & 0.3106  & 0.7088  & 0.7023  & 0.5757  \\
\hline
FedSM \cite{fedsm2022}& 0.6894  & 0.6278  & 0.8021  & 0.7391  & 0.7146  \\
 LC-Fed \cite{lcfed}& 0.6233  & 0.4982  & 0.8217  & 0.7654  & 0.6772  \\
\hline
Only Local Training & 0.7049  & 0.4906  & 0.8079  & 0.7555  & 0.6897  \\
\hline
 \model{} (Ours)  & \textbf{0.7525}  & \textbf{0.7010}  & \textbf{0.8469}  & \textbf{0.7769}  & \textbf{0.7693}  \\
\hline
\end{tabular}%
\label{tab_seg}%
}
\vspace{6pt}
\captionof{table}{Ablation studies of \model{}.}
\vspace{-10pt}
\resizebox{1\linewidth}{!}{
\begin{tabular}{c|cc|c}
\hline
\multirow{2}[1]{*}{Methods} & \multicolumn{2}{c|}{Breast Cancer} & Segmentation \\
\cline{2-4}          & ACC$\uparrow$   & MF1$\uparrow$   & Dice$\uparrow$ \\
\hline
\model{} & \textbf{0.9038} & \textbf{0.8419} & \textbf{0.7693} \\
\hline
w/o Global Head & 0.8821 & 0.8192 & 0.7442 \\
w/o Global Body  & 0.8295 & 0.7366 & 0.7198 \\
w/o Features Weighted Fusion & 0.8588 & 0.8009 & 0.7455 \\
\hline
Only Local Training & 0.7875 & 0.7005 & 0.6897 \\
\hline
\end{tabular}%
\label{table_ab}%
}

\end{minipage}
\begin{minipage}[c]{0.53\textwidth}
\centering
 \captionof{table}{GFLOPS and parameters of local heterogeneous models and messenger models in various tasks. The smaller the better. Among the four tasks, the GFLOPS and parameters of the messenger models are much smaller than those of the local models.}
  \resizebox{1\linewidth}{!}{
    \begin{tabular}{c|c|ccc}
    \toprule
    \textbf{Tasks} & \textbf{Dataset} & \textbf{Model} & \textbf{GFLOPS} & \textbf{\#Params} \\
    \hline
    \multicolumn{1}{c|}{\multirow{5}[4]{*}{\makecell{Medical Image \\ Classification \\ (Different Resolution)}}} & \multicolumn{1}{c|}{\multirow{5}[4]{*}{\makecell{BreaKHis\\(384x384x3\\- 48x48x3)}}} & ResNet17 & 3.495  & 4.231M \\
          &       & ResNet11 & 0.667  & 2.104M \\
          &       & ResNet8 & 0.140  & 1.558M \\
          &       & ResNet5 & 0.044  & 1.359M \\
\cline{3-5}          &       & Messeger & \textbf{0.01-0.07} & \textbf{0.035M} \\
    \hline
    \multicolumn{1}{c|}{\multirow{18}[8]{*}{\makecell{Medical Image \\ Classification \\ (Different Label \\ Distributions)}}} & \multicolumn{1}{c|}{\multirow{9}[4]{*}{\makecell{BreaKHis\\(384x384x3)}}} & ResNet & 10.020  & 11.111M \\
          &       & Shufflenetv2 & 1.719  & 1.730M \\
          &       & ResNeXt & 41.245  & 7.930M \\
          &       & squeezeNet & 7.774  & 1.832M \\
          &       & SENet & 80.370  & 12.372M \\
          &       & MobileNet & 1.870  & 1.934M \\
          &       & DenseNet & 13.461  & 1.147M \\
          &       & VGG   & 57.524  & 40.045M \\
\cline{3-5}          &       & Messeger & \textbf{0.070} & \textbf{0.032M} \\
\cline{2-5}          & \multicolumn{1}{c|}{\multirow{9}[4]{*}{\makecell{OCT 2017\\(256x256x1)}}} & ResNet & 4.351  & 11.090M \\
          &       & Shufflenetv2 & 0.735  & 1.712M \\
          &       & ResNeXt & 18.256  & 7.910M \\
          &       & squeezeNet & 3.342  & 1.820M \\
          &       & SENet & 35.644  & 12.363M \\
          &       & MobileNet & 0.812  & 1.921M \\
          &       & DenseNet & 5.954  & 1.14M \\
          &       & VGG   & 25.501  & 40.020M \\
\cline{3-5}          &       & Messeger  & \textbf{0.012 } & \textbf{0.035M} \\
    \hline
    \multicolumn{1}{c|}{\multirow{5}[4]{*}{\makecell{Medical Image \\ Segmentation Task}}} & \multicolumn{1}{c|}{\multirow{5}[4]{*}{\makecell{ColonDB \\ETIS \\ClinicDB \\Kvasir-SEG \\ (256x256x3) \\ \ }}} & Unet++ & 34.906  & 10.421M \\
          &       & FCN   & 54.742  & 32.560M \\
          &       & Unet  & 56.435  & 33.090M \\
          &       & ResUnet & 25.572  & 19.913M \\
\cline{3-5}          &       & Messeger  & \textbf{0.681} & \textbf{0.196M} \\
    \bottomrule
    \end{tabular}%
    }
  
\end{minipage}

\end{table}


\subsection{Ablation Experiments}
To verify the effectiveness of \model{}'s key components, we conduct a comparative analysis by removing each of the three elements (global head, global body, and feature-weighted fusion) during breast cancer classification tasks with different label distributions and segmentation tasks, as shown in Table 4. 
The experimental results indicate that more parameter sharing is beneficial for \model{}, and features weighted fusion effectively improves the performance of local heterogeneous models. 
 
\subsection{GFLOPS and Parameters}
We compare the GFLOPS and parameter of the global bypass model with local heterogeneous models in three tasks. The results of Table 5 show that the GFLOPS and parameters of the global bypass model are much smaller than those of the local heterogeneous model on all the tasks.

\section{Conclusion}

\model{} can effectively solve the problems of statistic heterogeneity and system heterogeneity faced in federated learning. \model{}, based on the global bypass model paradigm, offers a solution to these issues. \model{} introduces a lightweight global bypass model in each client and designs a feature weighted fusion to fuse local and global knowledge. These can enable local heterogeneous models to capture information from other clients well under statistic heterogeneity. Numerous experiments have demonstrated that our method outperforms existing federated learning frameworks with heterogeneous models in multiple tasks.

\subsubsection{Acknowledgements.} This work was supported by the National Key R$\&$D Program of China under Grant No.2022YFB2703301.

\subsubsection{Disclosure of Interests.} The authors have no competing interests to declare that are relevant to the content of this article.

\end{document}